\documentclass[10pt,twocolumn,letterpaper]{article}

\usepackage[pagenumbers]{cvpr} 

\usepackage[dvipsnames]{xcolor}

\usepackage[accsupp]{axessibility} 

\definecolor{cvprblue}{rgb}{0.21,0.49,0.74}
\usepackage[pagebackref,breaklinks,colorlinks,citecolor=cvprblue]{hyperref}

\def\paperID{55} 
\def\CodalabID{Steve} 
\def\confName{CVPR}
\def\confYear{2024}

\title{Learnable Prompt for Few-Shot Semantic Segmentation\\ in Remote Sensing Domain}

\author{
Steve Andreas Immanuel\\
TelePIX\\
Seoul, South Korea\\
{\tt\small steve@telepix.net}
\and
Hagai Raja Sinulingga\\
TelePIX\\
Seoul, South Korea\\
{\tt\small hagairaja@telepix.net}
}

\begin{document}
\maketitle
\begin{abstract}
Few-shot segmentation is a task to segment objects or regions of novel classes within an image given only a few annotated examples. In the generalized setting, the task extends to segment both the base and the novel classes. The main challenge is how to train the model such that the addition of novel classes does not hurt the base classes performance, also known as catastrophic forgetting. 
To mitigate this issue, we use SegGPT as our base model and train it on the base classes. Then, we use separate learnable prompts to handle predictions for each novel class. To handle various object sizes which typically present in remote sensing domain, we perform patch-based prediction. To address the discontinuities along patch boundaries, we propose a patch-and-stitch technique by re-framing the problem as an image inpainting task. During inference, we also utilize image similarity search over image embeddings for prompt selection and novel class filtering to reduce false positive predictions. Based on our experiments, our proposed method boosts the weighted mIoU of a simple fine-tuned SegGPT from 15.96 to 35.08 on the validation set of few-shot OpenEarthMap dataset given in the challenge.
\end{abstract}    
\section{Introduction}
\label{sec:intro}

Generalized Few-Shot Segmentation (GFSS) is a computer vision task wherein the model must effectively segment novel classes with limited examples alongside the base classes it has been trained on. This task holds significant relevance in remote sensing applications, where annotation costs are high and diverse user interests necessitate adaptable segmentation models. For instance, agricultural societies prioritize distinguishing between cultivated and fallow land, whereas civil registration departments require accurate house mapping for population estimation. Recognizing the critical importance of addressing these challenges, the OpenEarthMap Land Cover Mapping Few-Shot Challenge \cite{xia_2023_openearthmap}, jointly organized with the \href{https://codalab.lisn.upsaclay.fr/competitions/17568}{L3D-IVU CVPR 2024 Workshop}, was convened to drive advancements in solving the problem.

The strategy for addressing the GFSS problem currently revolves around two approaches: 1) individually predicting each novel class and then merging the results using fusion techniques \cite{min2021hsnet, lang2022bam}, and 2) relearning the classifier so it can predict both base and novel classes simultaneously \cite{liu2023learning_pop, hajimiri2023_diam, tian2022gfsseg_capl}. Our method follows the first approach but diverges from existing methods in that we train the model only once using data from the base class. For the novel classes segmentation, we only derive a prompt for each class obtained by training solely on the support set. The reason we chose this is due to the emergence of new foundation models with strong generalization capabilities \cite{seggpt_2023, kirillov2023segment_SAM, clipvit}. The prompt for each novel class serves as an adaptation layer to handle a specific novel class characteristics. Therefore, our method is able to handle any number of novel classes without performance degradation on the base classes. This approach is both simple and straightforward, yet highly extensible to real-life scenarios.

Furthermore, this challenge presents characteristics commonly encountered in remote sensing, particularly varying object sizes. In remote sensing imagery, there are both large-scale objects like industrial complexes, roads, and lakes, as well as smaller objects such as trees, boats, and houses. Common strategies include employing multi-scale features \cite{cheng2023semantic_pyramid} to capture both large and small objects, or segmenting the image into smaller patches \cite{wang2023deep_patch}. Our approach incorporates elements of both strategies, utilizing a comprehensive foundational model comprising multi-scale layers and leveraging detailed inference results from smaller patches. Although segmenting images into patches is not widely favored due to the risk of information loss and discontinuous results along patch boundaries, our method addresses this issue by introducing a patch-and-stitch technique.

The contribution and novelty of our approach can be summarized as:
\begin{itemize}
\item We introduce a simple, yet effective method to handle novel classes prediction in few-shot setting using learnable prompts. Initial training is only done on the base classes while the learnable prompts are optimized using the frozen model. 
\item We propose a patch-and-stitch technique to smooth out the results in patch-based predictions, especially along the patch boundaries. We also incorporate similar prompt searching based on similarity and novel class filtering to further boost the performance.
\end{itemize}
\section{Related Work}
\label{sec:relatedwork}

\subsection{Semantic Segmentation}
The basis of GFSS is semantic segmentation where models assign labels to individual pixels in an image. While methods like FCN \cite{long2015FCN} and encoder-decoder architectures \cite{ronneberger2015unet, badrinarayanan2017segnet} have improved per-pixel predictions, incorporating context information through techniques like dilated convolutions and attention mechanisms has been proven to further enhance segmentation accuracy. Recent advancements, including pyramid pooling \cite{hou2020strip}, parallel dilated convolutions \cite{chen2018deeplbv}, and the adoption of vision transformers \cite{strudel2021segmenter}, have led to significant improvements in segmentation quality. However, challenges persist in adapting these models to handle unseen classes without extensive fine-tuning using sufficiently annotated data.

\subsection{Few-Shot Semantic Segmentation}
Few-shot semantic segmentation (FSS) is formulated to specifically answer the challenge wherein pixel-wise labeling is required for novel classes with limited support examples, mainly focusing on the novel class prediction only without the base class. Methods like PL \cite{dong2018few_PL} and PANet \cite{wang2019panet} adapt prototype learning, while ASR \cite{liu2021anti_ASR} learns orthogonal prototypes. Other method like OSLSM \cite{shaban2017oslsm} assigns weights to the final classifier and PFENet \cite{tian2020prior_PFENet} leverages pre-trained backbone knowledge and addresses spatial inconsistency with a Feature Enrichment Module (FEM). Despite FSS models' effectiveness with support samples, the practical scenarios involving both base and novel classes on the target image has not been covered by these methods.

\subsection{Generalized Few-Shot Semantic Segmentation}
In the generalized setting, the task is to predict not only the novel classes but also the base classes. Several approaches that have been proposed can be seperated into two categories based on the prediction design setting. \newline
\textbf{Directly predict all novel and base classes}
\begin{itemize}
    \item \textit{POP} \cite{liu2023learning_pop}, employs orthogonal loss to separate base and novel classes from the background. The training process involves two stages: initially training a PSPNet to obtain the base model, and then training the novel model using combined data from both base and novel classes, with the base data chosen through random sampling. Notably, the POP model updates labels from the base class at every epoch.
    \item \textit{DIaM} \cite{hajimiri2023_diam}, focuses on finetuning the classifier using a modified InfoMax framework and knowledge distillation techniques to retain base class knowledge.
    \item \textit{CAPL} \cite{tian2022gfsseg_capl}, pioneers GFSS with three classifiers: one for base class together with novel class, one for base class only, and a classifier to combine both result.
\end{itemize}
\textbf{Predict classes individually}
\begin{itemize}
    \item \textit{BAM} \cite{lang2022bam}, trains a ResNet50 encoder and decoder for each novel class, utilizing weights to select support images. However, this approach is tailored for the one-class novel scenario and the performance is reported to degrade as the number of novel classes increased \cite{hajimiri2023_diam}.
    \item \textit{HSNet} \cite{min2021hsnet}, utilizes a 4D sparse correlation tensor over feature pyramids generated by a CNN-based backbone network. Although HSNet is not designed for GFSS, but adaptation of the model as reported in \cite{liu2023learning_pop} show a competitive result.
\end{itemize}

\subsection{Foundation Model}
The concept of training large-scale models through semi-supervised learning on extensive datasets, referred to as an upstream task, and subsequently employing them as foundation models for fine-tuning on downstream tasks, has gained significant prominence recently \cite{painter_2023, clipvit}. These large models exhibit strong generalization capabilities due to their training on diverse datasets. Some of these models have been explored for addressing general remote sensing tasks such as segmentation, object detection, change detection and super resolution \cite{cong2022satmae_foundation, manas2021seasonal_foundation_seco, mendieta2023towards_foundation_continual, bastani2023satlaspretrain, akiva2022matter, mall202caco}. Notably, recent research \cite{seggpt_2023} demonstrates that even without fine-tuning on specialized remote-sensing data, this approach can rival SOTA FSS methods, particularly with an increasing number of shots. This observation motivate our direction in exploring the potential of the foundation model, as no prior attempts have been made to tackle the GFSS problem using such approach.
\section{Method}
\label{sec:method}

\subsection{Preliminaries}
Given an RGB image $\textbf{X} \in \mathbb{R}^{H \times W \times 3}$, where $H$ and $W$ denote the height and width of the image, the goal of semantic segmentation task is to predict semantic map $\textbf{Y} \in \mathbb{R}^{H \times W}$. Each pixel $\textbf{Y}_{ij}$ corresponds to a class label from a predefined set $\mathcal{C}=\{c_1, ..., c_P\}$, where $P$ is the number of classes, reflecting the semantic class of the corresponding pixel $\textbf{X}_{ij}$. In few-shot setting, the training dataset contains only base classes, while for each novel class, we are given $k$ samples of images and their semantic maps which only contain the novel class.

\subsection{Model Architecture}
We use SegGPT \cite{seggpt_2023} as the foundation model due to its strong generalizability. Internally, SegGPT uses ViT \cite{vit} as the backbone and is trained with smooth-$l_1$ loss.

\subsection{Training}
\begin{figure}
    \centering
  
    \begin{subfigure}{\linewidth}
        \centering
        \includegraphics[width=\linewidth]{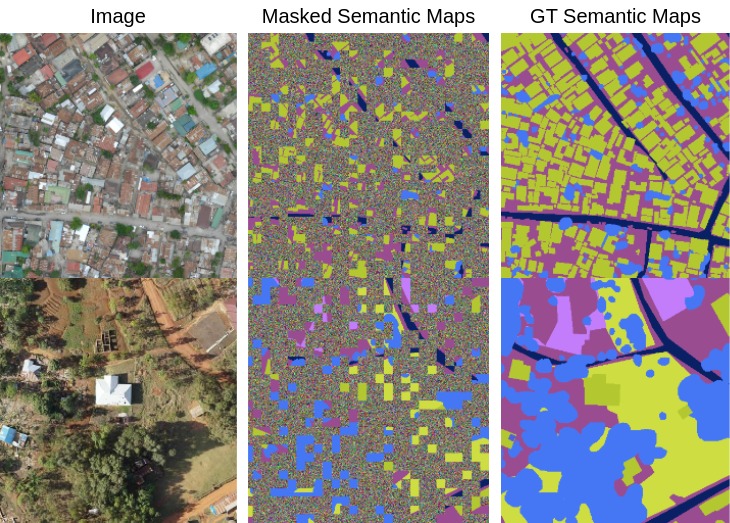}
        \caption{Random masking strategy}
        \label{fig:mim-a}
    \end{subfigure}

    \vspace{10pt}
  
    \begin{subfigure}{\linewidth}
        \centering
        \includegraphics[width=\linewidth]{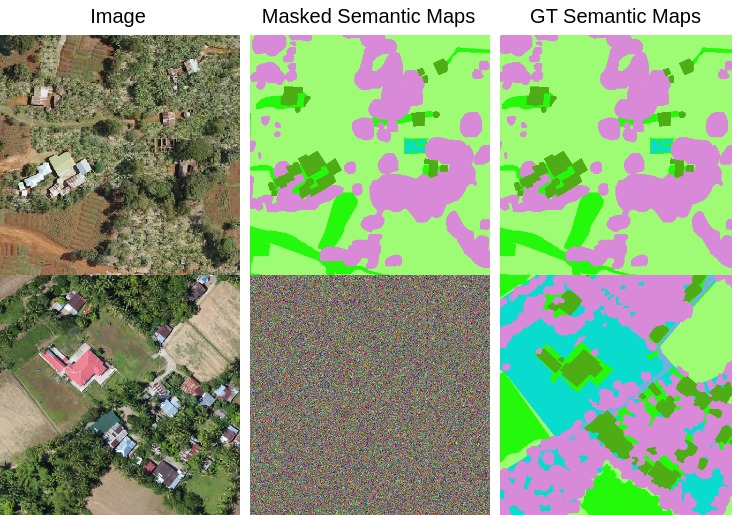}
        \caption{Half masking strategy}
        \label{fig:mim-b}
    \end{subfigure}
  
    \caption{Masked image modeling approach}
    \label{fig:mim}
\end{figure}

We follow the masked image modeling (MIM) approach where the objective is to reconstruct the masked regions of the input image. To this end, a pair of images is fed to the model instead of only one image. Prompt image $\textbf{X}^p$ and target image $\textbf{X}^t$, and their corresponding semantic maps $\textbf{Y}^p$ and $\textbf{Y}^t$, are provided, where certain patches of the semantic maps are masked as shown in \cref{fig:mim}. All $\textbf{X}^p$, $\textbf{X}^t$, $\textbf{Y}^p$, and $\textbf{Y}^t$ need to be of the same dimension $H$ x $W$ x 3. Therefore, the semantic maps are transformed into image space by mapping each class label into a color using color map $\mathcal{M}:\mathbb{R} \rightarrow \mathbb{R}^3$. This color is randomized for each data sample. The idea is to force the model to learn the contextual information in order to reconstruct the masked region, rather than exploiting the color \cite{seggpt_2023}. This is particularly useful in few-shot setting, because it prevents overfitting to the base classes.

\subsubsection{Base Classes}
The base classes are trained following the standard MIM approach. Each data sample consists of $\textbf{X}^p$, $\textbf{X}^t$, $\textbf{Y}^p$, and $\textbf{Y}^t$. In order to select $\textbf{X}^p$ and $\textbf{X}^t$, we initially generate all possible pair combinations of images in the training set. Then, we adopt different masking strategy for each pair depending on the classes present in each image. If $\textbf{X}^p$ and $\textbf{X}^t$ contain at least one different class, we randomly mask $\alpha$ portion of the patches of $\textbf{Y}^p$ and $\textbf{Y}^t$ as shown in \cref{fig:mim-a}. Alternatively, we mask the whole $\textbf{Y}^t$ as shown in \cref{fig:mim-b} if and only if $\textbf{X}^p$ and $\textbf{X}^t$ contain the exact same classes.

The idea is that if $\textbf{X}^p$ and $\textbf{X}^t$ contain the same classes, then given $\textbf{Y}^p$, the model should be able to predict the whole $\textbf{Y}^t$. In contrast, if their classes differ, the model should reconstruct the masked patches by leveraging contextual information from the unmasked regions.

\subsubsection{Novel Classes}
Due to the limited number of samples, the novel classes cannot be trained with the same approach as base classes. SegGPT inherently has strong few-shot capabilities by feeding the $k$ samples as the segmentation context. However, as we show in \cref{tab:ablation}, it still does not suffice for this challenge.

The primary obstacle in few-shot setting is how to make the model able to predict the novel classes given few samples, while simultaneously retaining the performance on the base classes. To this end, we use a learnable prompt $\textbf{Z}$ which acts as $\textbf{X}^p$ and $\textbf{Y}^p$. After we train the model on the base classes, we freeze the whole model and optimize only $\textbf{Z}$. Given there are $N$ novel classes, we create $\{\textbf{Z}^i\}_{i=1}^N$, each tailored to the characteristics of the corresponding $i$-th novel class, and train them independently using samples from each class. The main strength of this approach is that the introduction of novel classes does not compromise the performance of the base classes. Moreover, the optimization for $\textbf{Z}$ using the $k$ samples is much faster than the initial training on the base classes. Every learnable prompt for each novel class only amounts to about 5MB of model parameters, which is highly practical. To predict the novel classes, we simply utilize the corresponding learned prompt $\textbf{Z}$ in a plug-and-play manner.

For novel classes training, we only use the half masking strategy. Since the semantic maps for the novel classes only contains the novel classes, this task reduces into a binary classification between the novel class and background. As the novel class is typically confined to small regions within the image, direct prediction of the whole image would make the model trivially predict everything as background. Therefore, the learnable prompt $\textbf{Z}$ is optimized in two phases.  In phase 1, we adopt a sliding window approach to crop the image into smaller patches, utilizing only those containing the novel class to optimize $\textbf{Z}$, thus facilitating the model's comprehension of the novel class. Subsequently, in phase 2, we incorporate all patches, including those featuring only background, to mitigate false positive predictions of the novel class. Additionally, we only use white color to represent the novel class and black color to represent background, as opposed to random color as in base classes training.

\subsection{Inference}
Inference works similarly as half masking strategy in training. The image $\textbf{X}^p$ and its semantic map $\textbf{Y}^p$ acts as the prompt to give contextual information. Then, given the target image $\textbf{X}^t$, the model predicts $\hat{\textbf{Y}}^t$. We generate a fixed color map $\mathcal{M}$ and use it to transform $\textbf{Y}^p$ into image space, and its inverse $\mathcal{M}^{-1}$ to transform the prediction $\hat{\textbf{Y}}^t$ back to class label space. The class of the $ij$-th pixel in $\hat{\textbf{Y}}^t$ can be determined as follows,
\begin{equation}
  \label{eq:convert_cmap}
  class(\hat{\textbf{Y}}^t_{ij}) = \arg\underset{c \in \mathcal{C}}{\min}\ d(\hat{\textbf{Y}}^t_{ij}, \mathcal{M}^{-1}(c)),
\end{equation}
where $c$ iterates over the set of class labels $\mathcal{C}$, and $d$ is cosine similarity distance.

$\textbf{Image Similarity Search}$. The quality of the prompt $\textbf{X}^p$ and $\textbf{Y}^p$ greatly affects the prediction result $\hat{\textbf{Y}}^t$. In general, the more similar $\hat{\textbf{X}}^p$ and $\hat{\textbf{X}}^t$, the better the result. Additionally, SegGPT can incorporate multiple prompts in order to generate more accurate results. We leverage  CLIP-ViT \cite{clipvit} to extract the embeddings of each images in the training set. Then, we retrieve the top-$l$ most similar images to $\textbf{X}^t$ using cosine similarity and use them as the prompt.

\begin{figure}
    \centering
    \includegraphics[width=\linewidth]{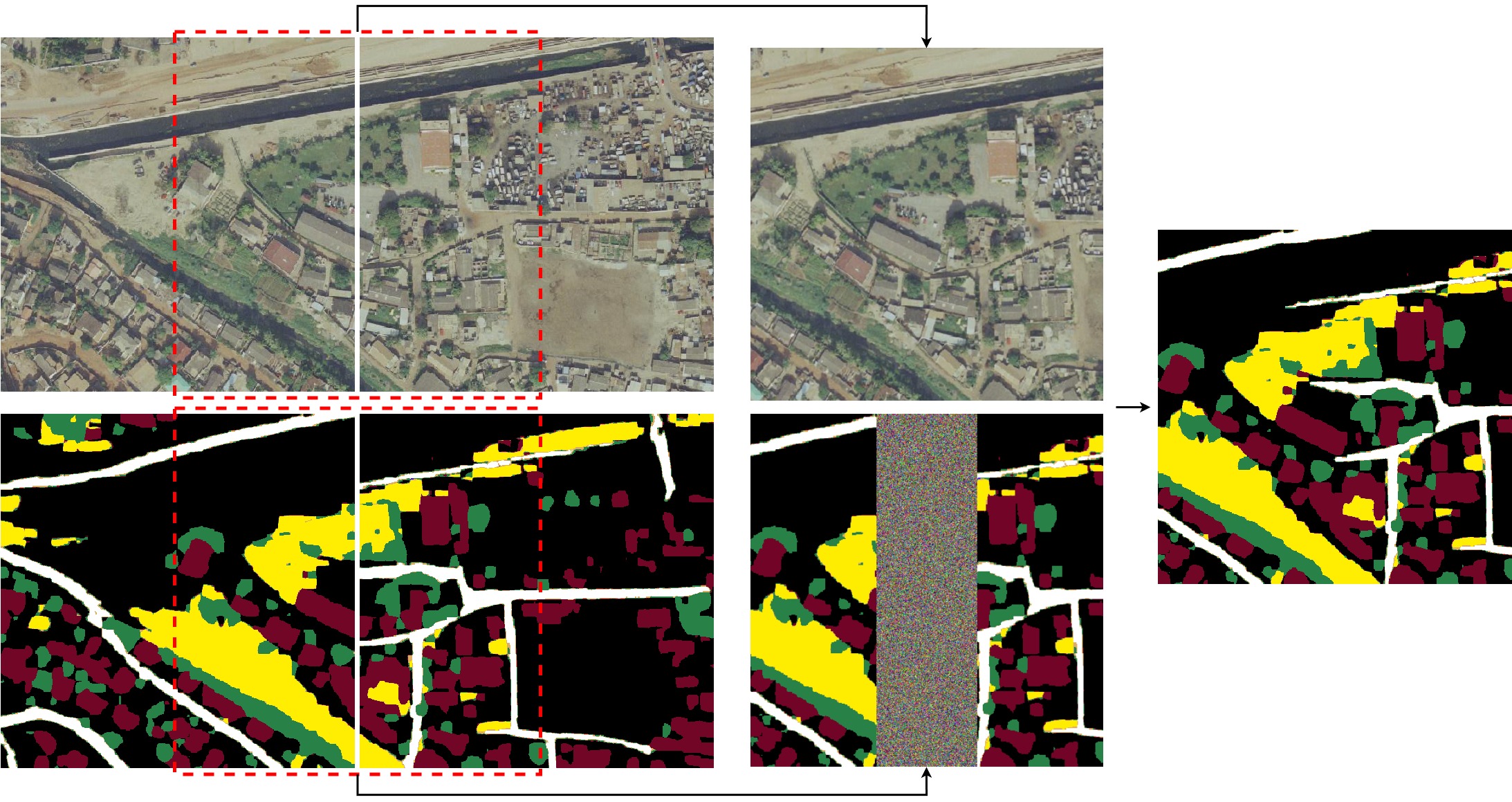}
    \caption{Seamless stitching between non-overlapping patches}
    \label{fig:stitch}
\end{figure}

\begin{figure*}
    \centering
    \includegraphics[width=0.8\textwidth]{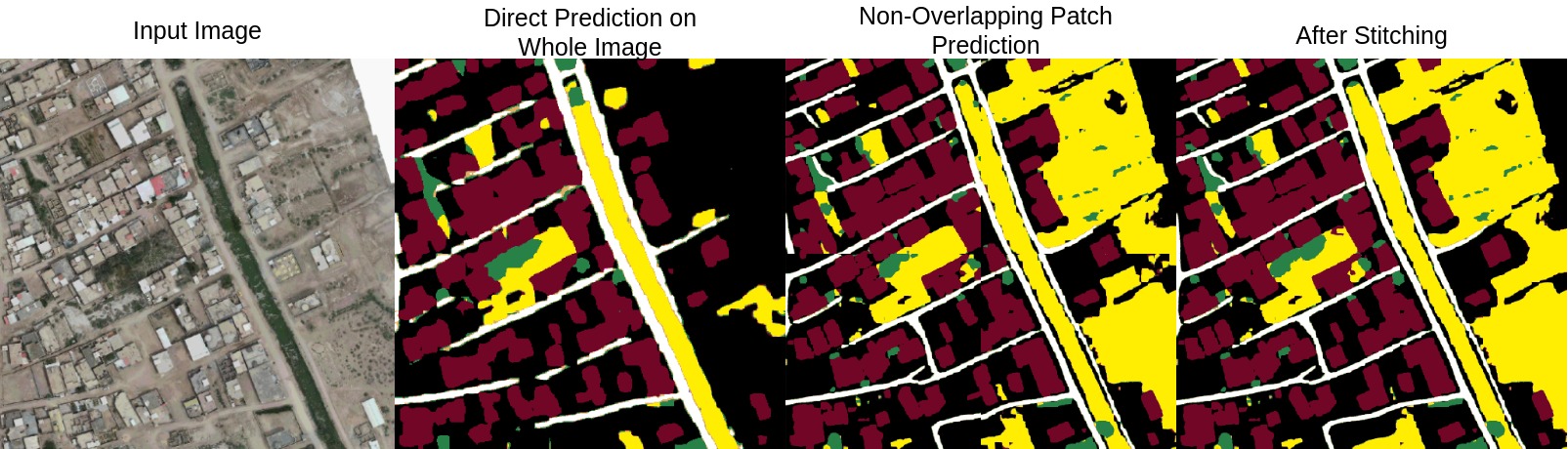}
    \caption{Seamless stitching produces more detailed and continuous result}
    \label{fig:qual-compare}
\end{figure*}

$\textbf{Patch-and-Stitch}$. In remote sensing domain, the objects are typically small and scattered across the image. Processing the whole image directly often leads to objects not being detected. Therefore, we partition the image into $2$x$2$ equal non-overlapping patches and perform the prediction on those patches independently. To get the result for the whole image, we can simply combine the prediction result on those patches directly. However, there might be some artefacts along the edges of the patches shown by the discontinuity of color (see \cref{fig:qual-compare} column 3). To mitigate this, we perform additional predictions on the middle regions that overlap adjacent patches as illustrated in \cref{fig:stitch}. Instead of predicting the entire overlapping region, we focus solely on predicting the middle portion, while the remaining areas are filled using the previous predictions from the non-overlapping patches to give more context. This process effectively frames the task as an image inpainting task, enabling seamless integration of non-overlapping patch predictions.

To get the final prediction containing both base and novel classes, we first perform prediction for the base classes. For each of the novel class, we do not utilize the image similarity search to get similar images as the prompt is essentially replaced with $\textbf{Z}$. Instead, we calculate the similarity between the target image $\textbf{X}^t$ and the $k$ given samples, analogously using CLIP-ViT and cosine similarity distance. If the similarity does not exceed a certain threshold we skip processing the corresponding novel class for the target image altogether. The idea is if $\textbf{X}^t$ is not similar with the $k$ given samples, then it is unlikely to contain the novel class. This approach helps to further reduce false positive prediction of novel classes. Subsequently, we simply overlay the novel classes predictions on top of the base classes prediction.

\section{Experiment}
\label{sec:experiment}

\begin{figure*}
    \centering
    \includegraphics[width=\textwidth]{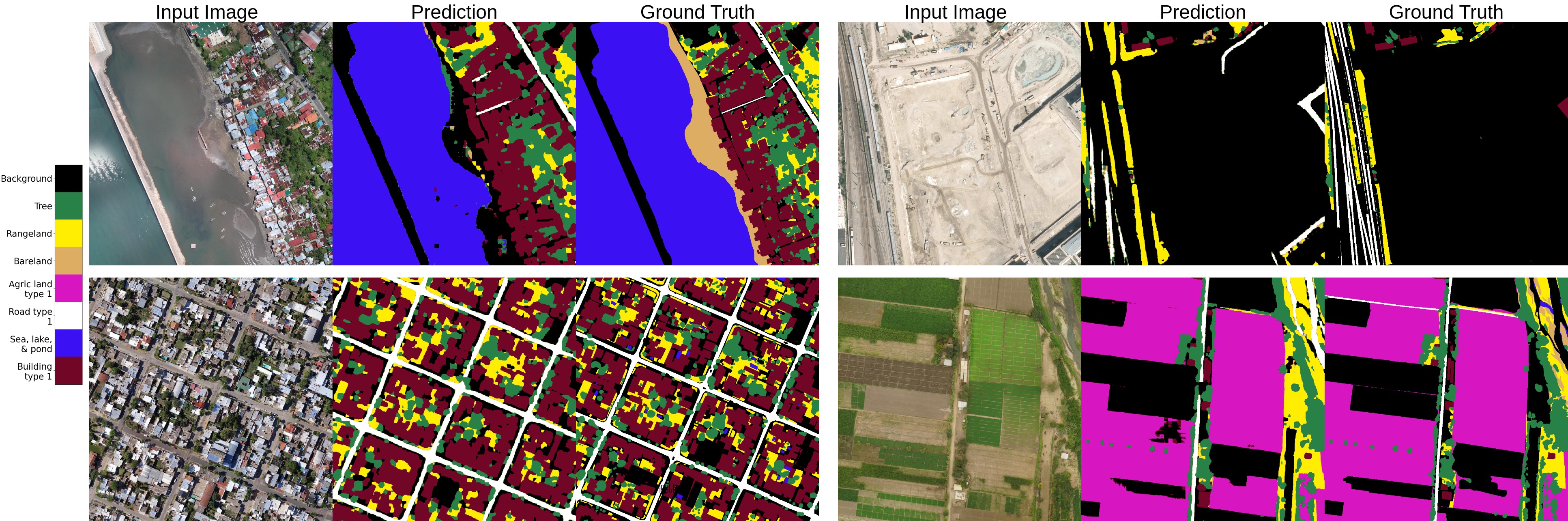}
    \caption{Semantic map prediction results on the training set}
    \label{fig:qual}
\end{figure*}

\subsection{Dataset}
The dataset used in the challenge is a few-shot dataset consists of 408 samples of the original OpenEarthMap (OEM) benchmark dataset \cite{xia_2023_openearthmap}. The challenge dataset extends the original 7 semantic classes (excluding background class) of the OEM to 15 classes, which is split into 7:4:4 for base classes, validation novel classes, and test novel classes, respectively.  All base, validation novel, and test novel classes are disjointed.

The 408 samples are split into 258 as training set, 50 as validation set, and 100 as test set. Validation and test set only contains novel classes. For each novel class, 5 example images are given with their corresponding semantic maps. Therefore, 20 images are given as the support set, while 30 and 80 images are used as the query set for the validation and test set, respectively.

\subsection{Implementation Details}
We initialize SegGPT with the pretrained checkpoint provided by the original authors \cite{seggpt_2023}. Other initial foundation model checkpoints that are specialized on remote sensing domain \cite{graft, cong2022satmae_foundation, mendieta2023towards_foundation_continual, bastani2023satlaspretrain, akiva2022matter, mall202caco} can also be used, which might improve the performance. We leave this for future work. During training, we use image augmentations including, random cropping, random horizontal and vertical flip, and color jittering. We set the $\alpha$ value for the random masking to 0.75 empirically. During inference, we select the top-5 most similar images as the prompt. We use AdamW \cite{adamw} optimizer and cosine annealing rate scheduler \cite{sgdr} with learning rate of 1e-4 and linear warmup for 500 steps by default. To optimize the learnable prompt, we simply treat $\textbf{Z}$ as model parameters of size $\mathbb{R}^{H \times W \times 3}$ times two (to represent $\textbf{X}^p$ and $\textbf{Y}^p$). All codes are implemented in Python using PyTorch. Training and experiments are conducted using 4 Nvidia RTX A6000.

\subsection{Evaluation Metric}
The evaluation metric is weighted mean intersection-over-union (mIoU) over all classes excluding the background. As the focus of the challenge is in the novel class, the final evaluation metric is calculated as $0.4 * \text{base classes mIoU} + 0.6 * \text{novel classes mIoU}$.

\subsection{Quantitative Results}
The mIoU results\footnote{The IoU for each class is not available because the detailed evaluation for some of our submissions cannot be viewed in the submission portal.} on the validation set of our proposed method are presented in \cref{tab:ablation}. We can see incremental improvements for each of the method that we employ. 

Simply using SegGPT that is finetuned on the OEM dataset only gives out mIoU of 15.96, mainly because the model is unable to detect any of the novel classes. Utilizing similar images as the prompt offers small improvement of +1.86 because it only improves the mIoU on the base classes. It is only after the integration of the learnable prompt that the model becomes capable of predicting the novel classes, leading to a substantial increase of mIoU by +7.44.

Incorporating the patch and stitch approach further enhances the mIoU to 29.41, demonstrating a significant improvement in capturing finer details and increasing overall accuracy. Filtering the novel classes based on the image similarity to the given $k$ samples also proves to be very effective, shown by another +5.67 increase in mIoU. By filtering the novel classes, we reduce the false positive prediction. Considering that we overlay the predictions of novel classes on top of the base classes, false positive predictions negatively impact not only the mIoU of the novel classes but also that of the base classes. As for the test set, we obtain a weighted mIoU of 36.52\footnote{The breakdown of each method is not available for the test set as in \cref{tab:ablation} due to the limited number of submissions during the competition and the submission portal is closed once the competition ends}.

\begin{table}
    \centering
    \caption{The impact of each method used on the mIoU of the validation set }
    \begin{tabular}{|l|l|}
        \hline
         Method & mIoU \\
        \hline 
        SegGPT Baseline \cite{seggpt_2023} & 15.96 \\
        + Similar image prompts & 17.82 \\
        + Learnable prompt on novel classes & 25.26\\
        + Patch and stitch & 29.41\\
        + Filter on novel classes & 35.08 \\
    \hline
    \end{tabular}
    \label{tab:ablation}
\end{table}

\subsection{Qualitative Results}
\cref{fig:qual} shows the results of our model on the training set. Our model ables to predict well on most classes. One particular class that our model struggle with is bareland, as shown in the top-left image in \cref{fig:qual}. This is mainly due to the very limited number of samples in the training set, as well as the inconsistency of the class definitions \cite{xia_2023_openearthmap}.

In \cref{fig:qual-compare}, we compare the results when directly predicting the whole image and using the patch and stitch method that we proposed. By using non-overlapping patch prediction, the model can capture finer details in the segmentation between buildings. Finally, the stitching mechanism allows aggregation of predictions from each patch seamlessly.

\subsection{Impact of Color Map on Inference}
While we can use any random color to fill $\mathcal{M}$, we observe empirically that some colors lead to higher performance. Specifically, we want objects that are typically located close together to have as different color as possible, e.g. bareland and sea, tree and road. Colors that are far apart from each other in RGB color space make it easier for the model to separate the close pixels between objects, reducing ambiguity. This is also similar to humans who can easily distinguish two adjacent objects that have contrasting colors.

\subsection{Limitations}
Due to how inference works in SegGPT, the results highly depend on the prompt that is given, therefore finding the most suitable prompt given an image is a crucial aspect to consider. The patch-and-stitch approach that we use in the inference also introduce additional computational cost as a trade-off for details and accuracy on the results.

\section{Conclusion}
\label{sec:conclusion}

In this work, we proposed a method to handle novel classes prediction in few-shot setting using learnable prompt. We also introduced some additional techniques including prompt based on image similarity, patch-and-stitch, and novel class filtering which led to substantial performance improvements.
{
    \small
    \bibliographystyle{ieeenat_fullname}
    \bibliography{main}
}


\end{document}